\documentclass[runningheads]{llncs}
\usepackage{graphicx}
\usepackage{caption}
\usepackage{subcaption}

\begin{document}
\title{Comparing Facial Expression Recognition in Humans and Machines: Using CAM, GradCAM, and Extremal Perturbation}
\titlerunning{Comparing Human and Machine FER}
%

\author{Serin Park\inst{1}\orcidID{0000-0001-7704-1167} \newline
Christian Wallraven\inst{2,*}\orcidID{0000-0002-2604-9115}}

\institute{Department of Artificial Intelligence, Korea University, Seoul, Korea \newline
\email{bvcxz565@korea.ac.kr}\newline
Department of Artificial Intelligence \& Department of Brain and Cognitive Engineering, Korea University, Seoul, Korea \newline
\email{wallraven@korea.ac.kr}
}

\authorrunning{Park and Wallraven}
\maketitle              
\begin{abstract}
Facial expression recognition (FER) is a topic attracting significant research in both psychology and machine learning with a wide range of applications. Despite a wealth of research on human FER and considerable progress in computational FER made possible by deep neural networks (DNNs), comparatively less work has been done on comparing the degree to which DNNs may be comparable to human performance. In this work, we compared the recognition performance and attention patterns of humans and machines during a two-alternative forced-choice FER task. Human attention was here gathered through click data that progressively uncovered a face, whereas model attention was obtained using three different popular techniques from explainable AI: CAM, GradCAM and Extremal Perturbation. In both cases, performance was gathered as percent correct. For this task, we found that humans outperformed machines quite significantly. In terms of attention patterns, we found that Extremal Perturbation had the best overall fit with the human attention map during the task.

\end{abstract}
\section{Introduction}
Facial expression is a natural and powerful tool of communication among humans. Facial expressions are instantly processed by humans conveying a wealth of messages: a smiling face spreads happiness, and a sad face makes our heart ache. There is evidence that some emotional facial expressions are largely universal \cite{ekman1997universal}, that is, they are recognized well in different cultures and associated with similar semantic content. Recent work, however, has cast some doubt to the degree of this universality \cite{gendron2014perceptions,jack2012facial} and the degree to which facial expressions are actually a reliable signal of an internal mental state \cite{barrett2019emotional}, but that does not lessen the importance of facial expressions in human-to-human communication \cite{nusseck2008contribution}.

Because of its importance to humans, facial expression recognition is also a major topic in the field of machine learning. If machines would be able to  interpret human facial expression correctly - and possibly make appropriate facial expressions in return - human-to-machine interaction would become more natural and efficient. Recognition of facial expression by machines is called automatic facial expression recognition, or automatic FER. Automatic FER has come a long way, from hand-crafted approaches to the current end-to-end deep learning models that locate and recognize facial expressions \cite{li2020deep}. Nonetheless, there still is a long way to go: current algorithms are good enough at recognizing laboratory-controlled facial expression images, but they struggle to recognize expressions from naturalistic images \cite{meng2019frame,savchenko2021facial}.

This leads us to the natural question: do humans and machines process facial expression images differently? And if they do, can we teach machines to act more like humans? In this paper, we adopted a human-in-the-loop (HIL) paradigm to address this question: we first collected click data from human participants to gather those spatial locations that may be important for disambiguating an expression in a two-alternative forced-choice task. Click data are reported to be a cost-efficient substitute for eye-tracking, and reflect the regional attention well \cite{egner2018attention,kim2017bubbleview}. For automatic FER, we trained an ensemble of deep neural networks on AffectNet \cite{mollahosseini2017affectnet}, a large, in-the-wild facial expression dataset, and compared its activation map with human click data using three different explainability or visualization methods. We also tried to further fine-tune the models with the human attention map to see whether this would improve the FER performance.

\section{Related work}
\subsection{Automatic FER}
After the advent of deep learning, FER has typically been implemented with deep neural networks due to their superior performance and robustness over the past years \cite{li2020deep}. In this section, we will go through some of the most popular datasets of FER and their benchmarks.

FER datasets can be broadly categorized as either controlled or in-the-wild. Controlled datasets are posed by trained actors and photographed in the lab with regular illumination. The extended Cohn-Kanade dataset (CK+; \cite{lucey2010extended}) is a classic example of a controlled dataset. It contains 593 video sequences from 123 individuals that start from neutral expression and culminate in the intended expression (one of seven categories: anger, contempt, disgust, fear, happiness, sadness, and surprise). In contrast, in-the-wild datasets are crawled from the web by searching for emotion-related keywords. This type of dataset is typically larger than controlled datasets, and noisier in terms of identity, illumination, etc. FER+ \cite{barsoum2016training}, a re-labeled version of FER2013 \cite{goodfellow2013challenges}, has been a popular early dataset in the field, with 28,709 training images, 3,589 validation images and 3,589 test images, consisting of the same seven expressions as the CK+ dataset. Currently, AffectNet \cite{mollahosseini2017affectnet} is the largest publicly-available labelled dataset on facial expressions (see Dataset section below for detailed information). 

FER models have reached excellent performance on controlled datasets: on CK+, Frame Attention networks proposed by \cite{meng2019frame} have attained 99.7\% accuracy. However, FER models perform less well with in-the-wild datasets: the state-of-the-art (SOTA) accuracy on the FER+ dataset with cleaned and updated labels is 89.75\% reached by a PSR model on seven expressions \cite{vo2020pyramid}. SOTA on the AffectNet dataset, in contrast, is only  65.74\% for seven of the eight included emotion categories \cite{savchenko2021facial}. Given that FER systems in practical use, such as humanoid robots and surveillance systems, will not be fed with regular illuminations, frontal head position and exemplary expressions, it is important to improve FER accuracy on such in-the-wild datasets.

\subsection{Human FER}
As automatic FER is an effort to mimic the natural capacity of humans to recognize facial expressions, one must look back on humans to get insight for the models. One thing to note is that humans are not necessarily better than deep neural networks in the task of classifying images. Human performance on the (non-updated) FER2013 dataset was 65$\pm$5\% \cite{goodfellow2013challenges}, while the then-SOTA model, ResMaskingNet \cite{pham2021facial} reached 76.82\%. However, this does not mean machines have outperformed humans in facial expression recognition in general. As stated above, recognizing someone's facial expression in the real world represents quite different challenges.

One of the key differences between humans and computers in FER is that humans pay attention to a limited region in the face, while computers treat all pixels equally in the initial phase. Humans distribute most of their attention to the eyes and mouth \cite{moon2019facial,nusseck2008contribution}, which partly explains the efficiency with which humans recognize facial expression. Interestingly, the region of interest can differ depending on culture \cite{jack2009cultural}. East Asians focus on the eyes, while Western Caucasian also pay attention to the mouth region. This difference leads East Asians to perform less well when discriminating 'fear' from 'disgust', in which pair the mouth region holds the key information.

\subsection{Transfer learning}
Transfer learning has two major approaches: the first approach is to pretrain and then fine-tune. Fine-tuning is a common practice in training neural networks. As it is difficult to collect large datasets for specific problems, researchers often train their networks first on ImageNet, a large-scale object classification dataset that contains 1.2 million images with 1000 classes \cite{deng2009imagenet}. ImageNet-pretrained models are also available in deep learning libraries. Fine-tuning may be done several times: \cite{ng2015deep} introduced cascaded fine-tuning, where the researchers first trained a deep CNN model on ImageNet dataset, an auxiliary dataset related to emotion recognition, and finally on the target dataset.

Another approach is Knowledge Distillation, also known as Teacher-Student model. It was originally developed as a model compression method \cite{bucilua2006model}. The Teacher network first learns the representations and outputs prediction labels. Then the Student network is trained on the prediction labels of Teacher. One key aspect of Knowledge Distillation is that it can transfer knowledge across models with different structure. It even enables human-machine transfer learning: \cite{schiller2020relevance} implemented this type of learning, albeit indirectly. The researchers first trained a Teacher model on an FER dataset and obtained a saliency map for each image by visualizing the activations of the Teacher model. Then the researchers masked the images by leaving only the most important parts of the image based on the saliency maps - usually around the eyes and mouth - and used the masked images to train the Student model. The masked images initially helped accelerate the training, but the acceleration was retained only if the training data was switched to unmasked images after a certain point - there was no effect on accuracy, however. The critical aspect was that the researchers validated the masked images by comparing them to eye-tracking results by human observers. As the saliency maps of Teacher model were shown to be similar to human attention maps, this work is an indirect example of human-machine transfer learning.

\section{Dataset}
We chose the AffectNet dataset \cite{mollahosseini2017affectnet} to train our models for two reasons. First, it is the largest labelled dataset for facial expressions, containing a total of 440,601 labelled images in eight categories: neutral, happy, sad, surprise, fear, disgust, anger, and contempt. Second, it seems to be a difficult dataset to improve on: it was collected by web-based crawling methods, and each image was labeled by one, though expert, human annotator. Therefore, the AffectNet dataset contains in-the-wild images that may be mislabelled or vague. Specifically, out of 36,000 images that were annotated by two human annotators to calculate agreement, their agreement was only 60.7 percent. Moreover, the dataset is highly imbalanced: the largest class, happy, contains 143,991 images while the smallest class, contempt, contains only 5,119 images. This imbalance reflects the real-world proportion of expression occurrences; one does observe happy expressions more often than one observes contempt.

The baseline for AffectNet was measured with AlexNet \cite{krizhevsky2012imagenet}. As the test set is not publicly available, the validation set, which contains 500 images for each expression, is used as the  benchmark dataset. Baseline accuracy with weighted loss was 58\% on the validation set. The state-of-the-art is an SL + SSL in-panting-pl model \cite{pourmirzaei2021using}, with an average accuracy of 61.72\% for eight emotion categories. Table \ref{tab1} summarizes major results on AffectNet benchmarks and the baseline. In cases where one paper listed several methods with slight differences on the ranking, we only chose one method with the best result. Moreover, we only list methods that were tested on all eight emotion categories, as we will focus also on eight categories in this paper.

\begin{table}
\centering
\caption{SOTA and baseline results on AffectNet dataset}\label{tab1}
\begin{tabular}{|l|l|l|}
\hline
Method & Accuracy & Reference \\
\hline
SL + SSL in-panting-pl (B0) & 61.72 & \cite{pourmirzaei2021using}\\ 
Distilled student &  61.60 & \cite{schoneveld2021leveraging}\\
Multi-task EfficientNet-B0 & 61.32 & \cite{savchenko2021facial}\\
RW loss & 61.03 & \cite{fan2020learning}\\
PSR & 60.68 & \cite{vo2020pyramid}\\
Baseline (Weighted Loss) & 58.0 & \cite{mollahosseini2017affectnet}\\
\hline
\end{tabular}
\end{table}

\begin{figure}
\begin{center}
\includegraphics[width=0.8\textwidth]{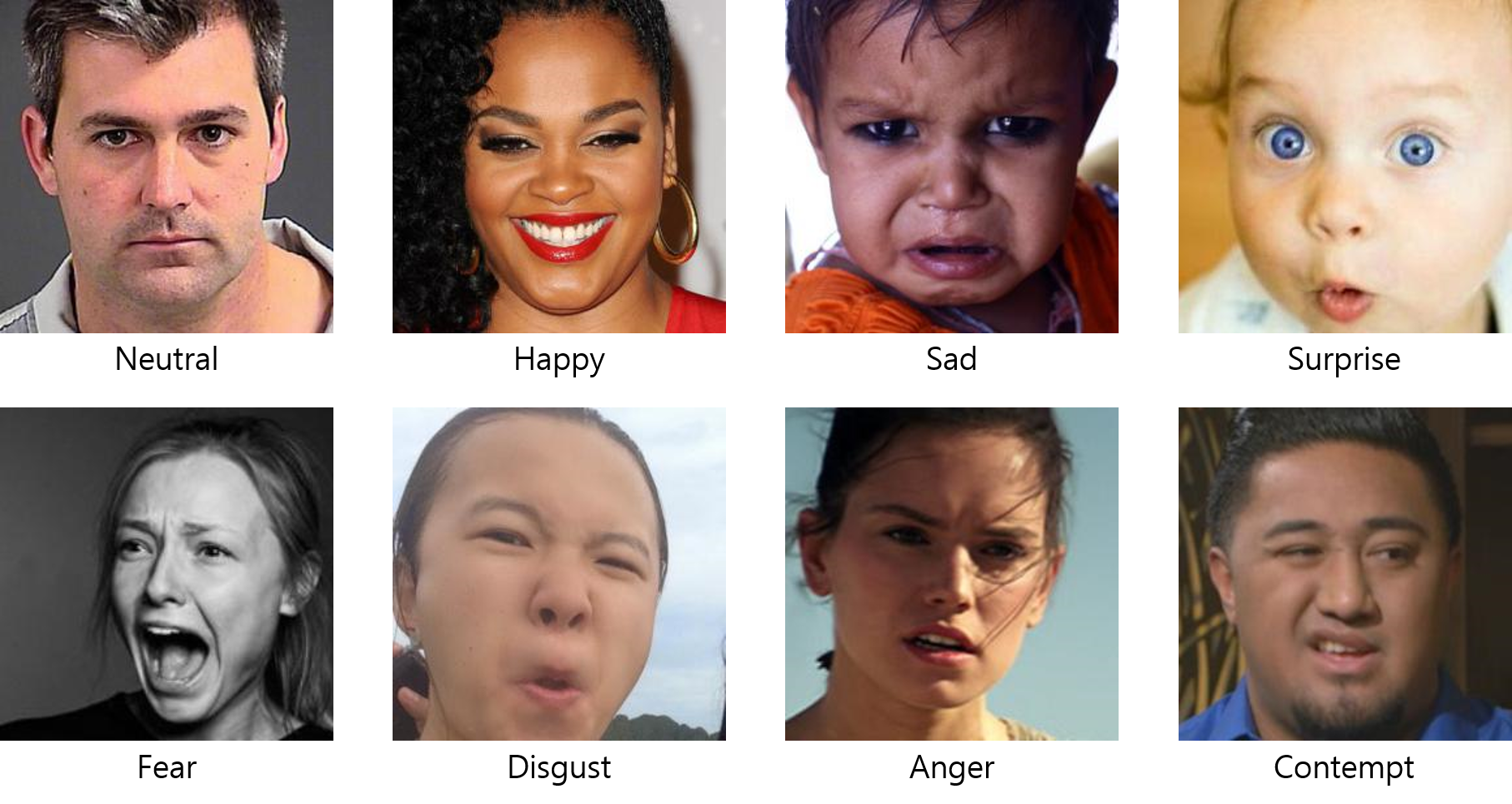}
\caption{Sample images used for human experiment} \label{fig:samples}
\end{center}
\vspace{-0.5cm}
\end{figure}

For the human experiment, hand-picked images from the AffectNet validation set were used (see Fig \ref{fig:samples} for example images). For each of the eight facial expression categories, 35 images were chosen that were deemed to be good representatives of the intended expressions. The total number of images was therefore 280. All computational experiments used the full training set of AffectNet and the validation set minus our 280 images.

\section{Human experiment}
\subsection{Participants}
24 Korea University undergraduates were recruited by an online advertisement (12 female, mean age $23.75\pm3.33$ years (SD) - all had normal or corrected-to-normal vision). Of the 24 participants, two had to be excluded from statistical analyses; one being an outlier (overall accuracy more than 3 $\sigma$ lower compared to the sample mean), and the other having skipped an entire block due to mistake. 

\subsection{Methods and task}
The 280 experimental images were blurred by an opencv function (cv2.blur) with $k=70$, and converted to grayscale, so that the expression could not be recognized by just looking at the blurred image. Human participants were asked to click on these blurred images to reveal circular parts. The revealed parts were not blurred and in original colors. From the revealed parts, participants had to determine which expression the picture portrayed. There was no limit to the number of clicks, but the participants were asked to make as few clicks as possible. Participants were seated in a quiet room in front of a monitor at a distance of roughly 57 cms. The faces subtended roughly 4.5$^{\circ}$ of visual angle.

The response options were given in a two-alternative forced-choice ($2\mathrm{AFC}$) paradigm, one being the correct label and the other the false label. For one picture, the false label was fixed across all participants, so there was a fixed picture set for a pair of labels; such as 'happy' versus 'sad'. Moreover, participants were instructed to pay attention to the option pair before clicking on the image and use that information to guide clicks. This instruction was given in order to find key regions for discriminating between a pair of expressions. For example, the mouth region is crucial for discriminating 'fear' from 'surprise' \cite{jack2009cultural}.
There was a total of 280 trials, as 35 images were selected from each of the 8 categories. Order of trial was randomized for each participant. The trials were split into 4 sessions of 70 trials each with breaks in-between. 

\subsection{Results}
\begin{figure}
\centering
\includegraphics[width=0.5\textwidth]{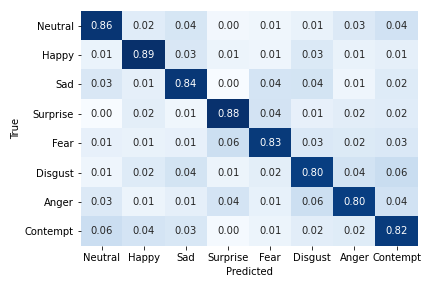}
\caption{Human Experiment confusion matrix} \label{fig:human_acc}
\end{figure}

The average accuracy across all participants  in this 2AFC task was 83.9\%. The confusion matrix (Figure \ref{fig:human_acc}) illustrates the response pattern of participants. The numbers in the cells are actual normalized values, but the colormap was based upon square roots of the values in order to highlight the differences among non-diagonal values. The confusion pattern shows that some pairs of emotions are confused more often than others: 'contempt' is mistaken for 'neutral', 'fear' for 'surprise', 'anger' for 'disgust', and 'disgust' for 'contempt'.

\begin{figure}
     \centering
     \begin{subfigure}[b]{0.32\textwidth}
         \centering
         \includegraphics[width=\textwidth]{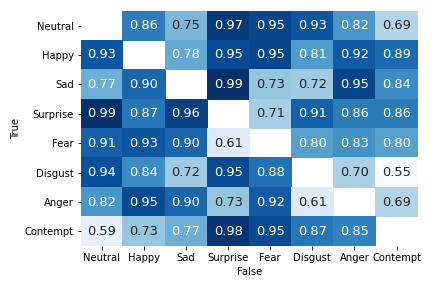}
         \caption{Accuracy}
         \label{fig:acc}
     \end{subfigure}
     \begin{subfigure}[b]{0.32\textwidth}
         \centering
         \includegraphics[width=\textwidth]{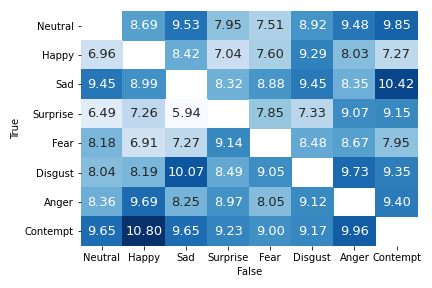}
         \caption{Number of clicks}
         \label{fig:count}
     \end{subfigure}
     \begin{subfigure}[b]{0.32\textwidth}
         \centering
         \includegraphics[width=\textwidth]{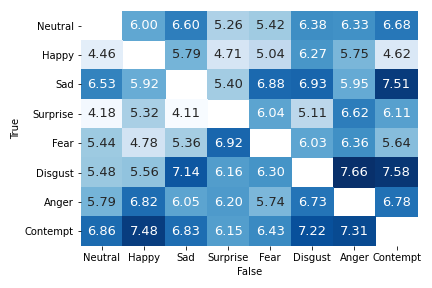}
         \caption{Time}
         \label{fig:time}
     \end{subfigure}
        \caption{Accuracy, number of clicks and time as a function of pairs}
        \label{fig:three cms}
\end{figure}

Next, we plotted the accuracy for each pair of expressions as a heatmap (Figure \ref{fig:acc}). 'True' labels (on the y axis) mean the true label for a given image, and 'false' labels (on the x axis) are the false options in the $2\mathrm{AFC}$ experiment. This matrix is not symmetric since an image with 'happy' as true label and 'sad' as false label is qualitatively different from an image with 'sad' as true label and 'happy' as false label.
Same-label pairs do not exist in the experiment, but were included as empty cells for a more legible visualization of pair structures. We also explored two other variables as a function of expression pairs: the number of clicks before label decision for one image, and the time taken between the first click and the label decision of one image (Figures \ref{fig:count},\ref{fig:time}). As in the accuracy heatmap, same-label pairs were included as empty cells. A clear positive correlation was observed between number of clicks and time, $r=.93$, $p<.001$. Significant negative correlations were observed between number of clicks and accuracy, $r=-.56$, $p<.001$, and between time and accuracy, $r=-.65$, $p<.001$.

\vspace{-0.7cm}

\begin{figure}
     \centering
     \begin{subfigure}[b]{0.48\textwidth}
         \centering
         \includegraphics[width=\textwidth]{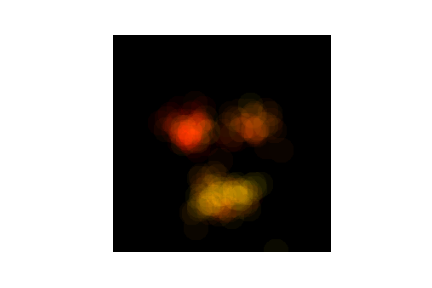}
         \caption{High accuracy (100\%)}
         \label{fig:click_seq_high}
     \end{subfigure}
     \begin{subfigure}[b]{0.48\textwidth}
         \centering
         \includegraphics[width=\textwidth]{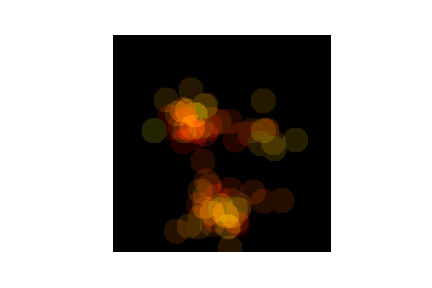}
         \caption{Low accuracy (32\%)}
         \label{fig:click_seq_low}
     \end{subfigure}
        \caption{Visualization of click sequence in images with high and low accuracy}
        \label{fig:click_seq}
\end{figure}

Next, we analyzed the pattern of clicks to investigate the strategies participants used. For visualization, clicks were color-coded according to their sequence: the first click was coded in red, the last click in yellow, and the clicks in between were given interpolated colors. We first obtained the colored click map for one picture, clicked by one participant, and averaged the click maps from all participants on the same picture, given that the label choice was correct. We found that participants mostly click the left eye first, then the right eye, and finally the mouth. In the case of images with low accuracy, the last few clicks were often around the left eye; which means people go back to the left eye when the stimulus is difficult to classify. Figure \ref{fig:click_seq_high} is an example of an image with high accuracy, where the trend to start from the left eye and to end at the mouth is clear. Figure \ref{fig:click_seq_low} is an example of an image with low accuracy, where the last click is often on the left eye.

\section{Algorithms}
\subsection{Model training}
To compare the human FER results with those of our DNN, the multiclass problem of classifying eight expression categories of the AffectNet dataset was split into a set of binary classification problems. Specifically, there were 28 classifiers (all possible pairs in eight emotions;${8 \choose 2}$) such as 'happy' versus 'sad', etc. For all classifiers, a ResNet-50 model was used with cross-entropy-loss and Adam optimizer with a base learning rate of 0.0001. Images were augmented by horizontal flip, small degrees of shifting, scaling, rotating, and changes in brightness and contrast. The batch size was 64. The classifiers were trained with original AffectNet training dataset restricted to the two categories in question, and the larger class was undersampled to match the smaller class. Therefore, the size of the dataset was different for each model. Because of this difference, all models were trained until they reached training accuracy of 90 (with maximum epoch of 150), rather than for a fixed number of epochs, to enable fair comparison. 

The trained binary classifiers were fine-tuned with the click-revealed images used in the human experiment. There were only 10 source images for each classifier, but as there were 22 participants, the size of fine-tuning dataset could be as large as 220, if the accuracy was 100 percent. The images were also augmented by shifting the click mask by one pixel in eight directions relative to the image. Lastly, we also gave unmasked versions of hand-picked images to the models to prevent catastrophic forgetting. These unmasked, or original, images were augmented in the same manner that was used in pretraining.

\subsection{Model results}
\begin{figure}
     \centering
     \begin{subfigure}[b]{0.48\textwidth}
         \centering
         \includegraphics[width=\textwidth]{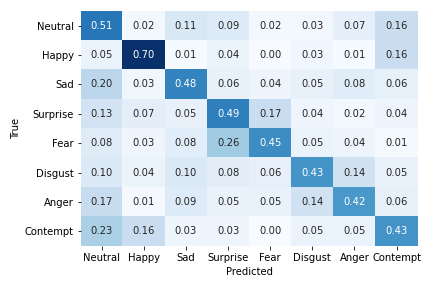}
         \caption{Simple Voting}
         \label{fig:ensemble_acc90_sv}
     \end{subfigure}
     \begin{subfigure}[b]{0.48\textwidth}
         \centering
         \includegraphics[width=\textwidth]{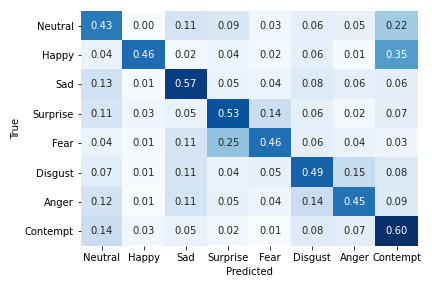}
         \caption{Weighted Voting}
         \label{fig:ensemble_acc90_wv}
     \end{subfigure}
        \caption{Ensembled results of pretrained models trained up to 90\% train accuracy}
        \label{fig:ensemble_acc90}
\end{figure}

We ensembled the prediction results from 28 binary classifiers by simple vote and weighted vote methods (Figure \ref{fig:ensemble_acc90_sv}, \ref{fig:ensemble_acc90_wv}). In the simple vote method, each classifier votes for a class for each test sample. The class with the most votes becomes the predicted label. In the weighted vote method, the largest output value from the fully connected layer of each classifier becomes the weighted vote \cite{galar2011overview}. This value reflects the confidence of the model. With the train-up-to-90 classifiers, accuracy was 49\% for simple vote and a similar level of 50\% for weighted vote.

\begin{figure}
     \centering
     \begin{subfigure}[b]{0.48\textwidth}
         \centering
         \includegraphics[width=\textwidth]{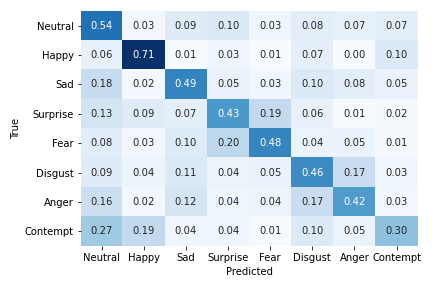}
         \caption{Simple Voting}
         \label{fig:ensemble_epoch30_sv}
     \end{subfigure}
     \begin{subfigure}[b]{0.48\textwidth}
         \centering
         \includegraphics[width=\textwidth]{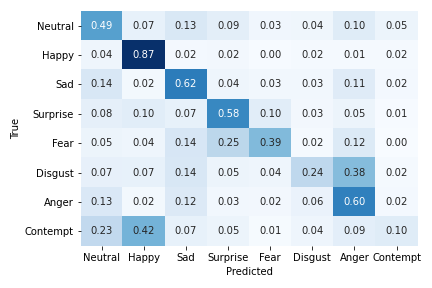}
         \caption{Weighted Voting}
         \label{fig:ensemble_epoch30_wv}
     \end{subfigure}
        \caption{Ensembled results of pretrained models trained for 30 epochs}
        \label{fig:ensemble_epoch30}
\end{figure}

Additionally, we tested a set of classifiers trained for 30 epochs to compare the pattern of prediction (Figure \ref{fig:ensemble_epoch30_sv}, \ref{fig:ensemble_epoch30_wv}). The performance was similar, with 48\% for simple vote and 49\% for weighted vote. However, classifiers trained for a set number of epochs showed a greater bias towards 'happy' in the confusion matrix. This tendency was more pronounced in the weighted vote method than in simple vote. This is because 'happy', the representation of which is learned in a relatively short time, gave high confidence while the confidence was low for other expressions.

\begin{figure}
\centering
\includegraphics[width=0.5\textwidth]{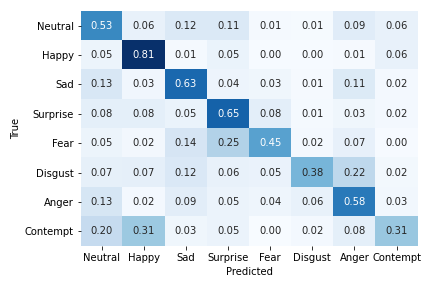}
\caption{Multiclass model for comparison} \label{fig:multiclass}
\end{figure}

We also trained a multiclass model for comparison (Figure \ref{fig:multiclass}). In this model, we implemented weighted cross entropy loss instead of undersampling. The number of epochs was 40, with other hyperparameters staying the same. The total accuracy of multiclass model was 54\%, which is higher than the ensemble model. However, the two minor classes, 'disgust' and 'contempt', showed higher recall rates for the ensemble model (Figure \ref{fig:ensemble_acc90_wv}). The pattern of confusion of the multiclass model was similar to Figure \ref{fig:ensemble_epoch30_sv}, where each pair is trained for the same number of epochs and the simple vote method is used. Overall, the ensemble method showed relatively similar performance across different classes, although the average accuracy dropped by 4\%.


\begin{figure}
     \centering
     \begin{subfigure}[b]{0.48\textwidth}
         \centering
         \includegraphics[width=\textwidth]{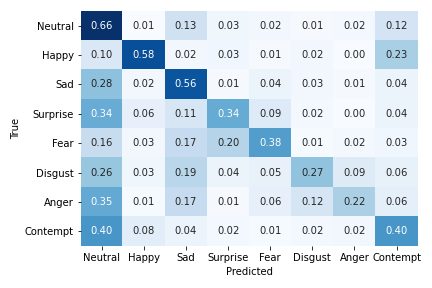}
         \caption{Simple Voting}
         \label{fig:finetuned_sv}
     \end{subfigure}
     \begin{subfigure}[b]{0.48\textwidth}
         \centering
         \includegraphics[width=\textwidth]{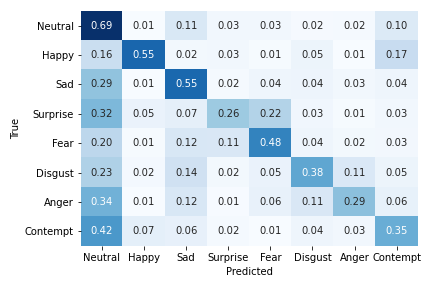}
         \caption{Weighted Voting}
         \label{fig:finetuned_wv}
     \end{subfigure}
        \caption{Ensembled results of finetuned models}
        \label{fig:finetuned}
\end{figure}

Lastly, we fine-tuned the models with masked images. Contrary to initial expectations, the overall accuracy \emph{decreased}, to 43\% for simple vote and 44\% for weighted vote. Varying the ratio of masked and unmasked images did not improve performance. In the confusion matrix (Figure \ref{fig:finetuned_sv}, \ref{fig:finetuned_wv}), we observed a strong bias towards the 'neutral' expression.

\subsection{Comparing humans and models}
We first looked at correlations of the confusion matrices for the different computational models and the human confusion matrix shown in Figure \ref{fig:human_acc}. There was a positive correlation between human and trained-up-to-accuracy-90 simple-vote model (Figure \ref{fig:ensemble_acc90_sv}), $r=.93$, $p<.001$, and weighted-vote model (Figure \ref{fig:ensemble_acc90_wv}), $r=.92$, $p<.001$. We also found a positive correlation between human and trained-for-30-epochs simple-vote model (Figure \ref{fig:ensemble_epoch30_sv}), $r=.91$, $p<.001$, and weighted-vote model (Figure \ref{fig:ensemble_acc90_wv}), $r=.80$, $p<.001$. Lastly, we found a positive correlation between human and multiclass model (Figure \ref{fig:multiclass}), $r=.91$, $p<.001$. This result supports our claim that although the ensemble model has lower average accuracy than multiclass model, its pattern of confusion is slightly more similar to that of humans than multiclass model.

\begin{figure}
     \centering
     \begin{subfigure}[b]{0.31\textwidth}
         \centering
         \includegraphics[width=\textwidth]{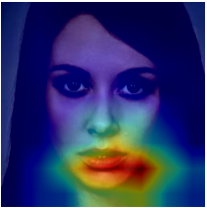}
         \caption{CAM}
         \label{fig:cam}
     \end{subfigure}
     \begin{subfigure}[b]{0.31\textwidth}
         \centering
         \includegraphics[width=\textwidth]{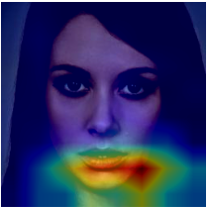}
         \caption{GradCAM}
         \label{fig:gradcam}
     \end{subfigure}
     \begin{subfigure}[b]{0.305\textwidth}
         \centering
         \includegraphics[width=\textwidth]{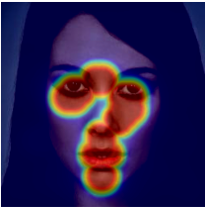}
         \caption{Extremal Perturbation}
         \label{fig:ep}
     \end{subfigure}
        \caption{Saliency maps for each method}
        \label{fig:three saliencies}
\end{figure}

We visualized the activations of pretrained models with three visualization techniques: CAM \cite{zhou2016learning}, GradCAM \cite{selvaraju2017grad} and Extremal Perturbation \cite{fong2019understanding}. Figure \ref{fig:three saliencies} demonstrates visualizations of each method over the same image.

To see how similar each method is to the human attention map, we computed dice coefficients between human attention maps and model saliency maps. The dice coefficient is given by two times the area of overlap between two binary masks, divided by the total number of 1's in both masks.
The attention, or saliency, maps were averaged within expression pairs, normalized and scaled to integer values between 0 and 255. The averaged masks were thresholded at value of 50; that is, values below 50 were set to 0 and values over 50 were set to 1.

\begin{figure}
\centering
\includegraphics[width=0.8\textwidth]{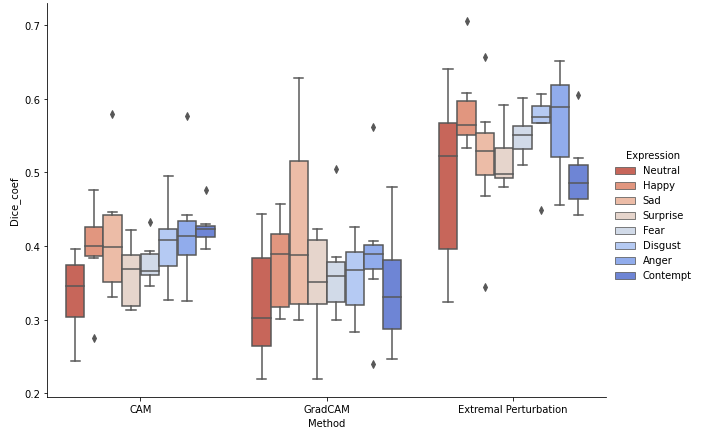}
\caption{Dice coefficients between human attention map and model saliency maps} \label{fig:dice}
\end{figure}

Figure \ref{fig:dice} illustrates dice coefficients of three different visualization methods in a box plot. If the dice coefficient is close to $1$, it means the method is similar to human attention maps. The plot reveals that Extremal Perturbation has the highest mean dice coefficient value. One-way analysis of variance (ANOVA) verifies this observation, as there is a highly significant difference among variables, $F=85.45$, $p<.001$. Pairwise Tukey analysis reveals that Extremal Perturbation had significantly higher dice coefficients than CAM, $t=10.31$, $p<.001$, and GradCAM, $t=12.12$, $p<.001$. There was no significant difference between CAM and GradCAM, $t=1.81$, $p > .05$. Additional analysis shows that the effect of facial expression was not significant, $F=1.34$, $p > .05$ and hence that all eight expressions were more similar at equivalent levels for Extremal Perturbation compared to the other two methods. 

\section{Conclusion and future work}

Our work compared human attention maps represented by click data, and model saliency maps using three different visualization methods: CAM, GradCAM, and Extremal Perturbation. We found that Extremal Perturbation had the best fit with the human attention. It will be interesting to extend this comparison also to other, more standard N-AFC FER tasks, in which humans need to disambiguate between more than two expressions.

Our computational experiments showed that the ensemble models of binary classifiers for FER did not perform as well as the standard multiclass model. However, by training the binary classifiers until they reach training accuracy of 90\% and combining the classification results by weighted vote method, we obtained a model that is less biased towards major expressions. Interestingly, we failed to improve the model using attentional information from humans as using the masked images as fine-tuning dataset proved an inadequate method for guiding the attention of a CNN-based model - this is in some way similar to the work by \cite{schiller2020relevance} who also found little to no improvement when using masked images. In the future, we will work on incorporating the attention mechanism to our model to channel the model's attention to meaningful regions more effectively.

Lastly, our participants pool was limited in that the participants were all East Asians. According to previous research \cite{jack2009cultural}, East Asians tend to focus only on the eyes compared to Western Caucasians, and are thus less accurate at discriminating between 'surprise' and 'fear', 'anger' and 'disgust', respectively. In Figure \ref{fig:click_seq}, we can actually see some evidence for this, with the mouth region often being the last to be revealed, which may be in line with this aforementioned research. Future experiments with Western Caucasian participants may show a different pattern of results.

\section{Acknowledgments}
This work was supported by Institute of Information Communications Technology Planning Evaluation (IITP; No. 2019-0-00079, Department of Artificial Intelligence, Korea University) and National Research Foundation of Korea (NRF; NRF-2017M3C7A1041824) grant funded by the Korean government (MSIT).

\bibliographystyle{splncs04}

{\small
\bibliography{ACPR}

\begin{thebibliography}{10}
\providecommand{\url}[1]{\texttt{#1}}
\providecommand{\urlprefix}{URL }
\providecommand{\doi}[1]{https://doi.org/#1}

\bibitem{barrett2019emotional}
Barrett, L.F., Adolphs, R., Marsella, S., Martinez, A.M., Pollak, S.D.:
  Emotional expressions reconsidered: Challenges to inferring emotion from
  human facial movements. Psychological science in the public interest
  \textbf{20}(1),  1--68 (2019)

\bibitem{barsoum2016training}
Barsoum, E., Zhang, C., Ferrer, C.C., Zhang, Z.: Training deep networks for
  facial expression recognition with crowd-sourced label distribution. In:
  Proceedings of the 18th ACM International Conference on Multimodal
  Interaction. pp. 279--283 (2016)

\bibitem{bucilua2006model}
Buciluǎ, C., Caruana, R., Niculescu-Mizil, A.: Model compression. In:
  Proceedings of the 12th ACM SIGKDD international conference on Knowledge
  discovery and data mining. pp. 535--541 (2006)

\bibitem{deng2009imagenet}
Deng, J., Dong, W., Socher, R., Li, L.J., Li, K., Fei-Fei, L.: Imagenet: A
  large-scale hierarchical image database. In: 2009 IEEE conference on computer
  vision and pattern recognition. pp. 248--255. Ieee (2009)

\bibitem{egner2018attention}
Egner, S., Reimann, S., Hoeger, R., Zangemeister, W.H.: Attention and
  information acquisition: Comparison of mouse-click with eye-movement
  attention tracking. Journal of Eye Movement Research  \textbf{11}(6),  1--27
  (2018)

\bibitem{ekman1997universal}
Ekman, P., Keltner, D.: Universal facial expressions of emotion. Segerstrale U,
  P. Molnar P, eds. Nonverbal communication: Where nature meets culture pp.
  27--46 (1997)

\bibitem{fan2020learning}
Fan, X., Deng, Z., Wang, K., Peng, X., Qiao, Y.: Learning discriminative
  representation for facial expression recognition from uncertainties. In: 2020
  IEEE International Conference on Image Processing (ICIP). pp. 903--907. IEEE
  (2020)

\bibitem{fong2019understanding}
Fong, R., Patrick, M., Vedaldi, A.: Understanding deep networks via extremal
  perturbations and smooth masks. In: Proceedings of the IEEE/CVF International
  Conference on Computer Vision. pp. 2950--2958 (2019)

\bibitem{galar2011overview}
Galar, M., Fern{\'a}ndez, A., Barrenechea, E., Bustince, H., Herrera, F.: An
  overview of ensemble methods for binary classifiers in multi-class problems:
  Experimental study on one-vs-one and one-vs-all schemes. Pattern Recognition
  \textbf{44}(8),  1761--1776 (2011)

\bibitem{gendron2014perceptions}
Gendron, M., Roberson, D., van~der Vyver, J.M., Barrett, L.F.: Perceptions of
  emotion from facial expressions are not culturally universal: evidence from a
  remote culture. Emotion  \textbf{14}(2), ~251 (2014)

\bibitem{goodfellow2013challenges}
Goodfellow, I.J., Erhan, D., Carrier, P.L., Courville, A., Mirza, M., Hamner,
  B., Cukierski, W., Tang, Y., Thaler, D., Lee, D.H., et~al.: Challenges in
  representation learning: A report on three machine learning contests. In:
  International conference on neural information processing. pp. 117--124.
  Springer (2013)

\bibitem{jack2009cultural}
Jack, R.E., Blais, C., Scheepers, C., Schyns, P.G., Caldara, R.: Cultural
  confusions show that facial expressions are not universal. Current biology
  \textbf{19}(18),  1543--1548 (2009)

\bibitem{jack2012facial}
Jack, R.E., Garrod, O.G., Yu, H., Caldara, R., Schyns, P.G.: Facial expressions
  of emotion are not culturally universal. Proceedings of the National Academy
  of Sciences  \textbf{109}(19),  7241--7244 (2012)

\bibitem{kim2017bubbleview}
Kim, N.W., Bylinskii, Z., Borkin, M.A., Gajos, K.Z., Oliva, A., Durand, F.,
  Pfister, H.: Bubbleview: an interface for crowdsourcing image importance maps
  and tracking visual attention. ACM Transactions on Computer-Human Interaction
  (TOCHI)  \textbf{24}(5),  1--40 (2017)

\bibitem{krizhevsky2012imagenet}
Krizhevsky, A., Sutskever, I., Hinton, G.E.: Imagenet classification with deep
  convolutional neural networks. Advances in neural information processing
  systems  \textbf{25},  1097--1105 (2012)

\bibitem{li2020deep}
Li, S., Deng, W.: Deep facial expression recognition: A survey. IEEE
  Transactions on Affective Computing  (2020)

\bibitem{lucey2010extended}
Lucey, P., Cohn, J.F., Kanade, T., Saragih, J., Ambadar, Z., Matthews, I.: The
  extended cohn-kanade dataset (ck+): A complete dataset for action unit and
  emotion-specified expression. In: 2010 ieee computer society conference on
  computer vision and pattern recognition-workshops. pp. 94--101. IEEE (2010)

\bibitem{meng2019frame}
Meng, D., Peng, X., Wang, K., Qiao, Y.: Frame attention networks for facial
  expression recognition in videos. In: 2019 IEEE International Conference on
  Image Processing (ICIP). pp. 3866--3870. IEEE (2019)

\bibitem{mollahosseini2017affectnet}
Mollahosseini, A., Hasani, B., Mahoor, M.H.: Affectnet: A database for facial
  expression, valence, and arousal computing in the wild. IEEE Transactions on
  Affective Computing  \textbf{10}(1),  18--31 (2017)

\bibitem{moon2019facial}
Moon, H.J.: Facial Expression Processing with Deep Neural Networks: from
  Implementation to Comparison with Humans. Master's thesis, Korea University,
  Seoul, Korea (2019)

\bibitem{ng2015deep}
Ng, H.W., Nguyen, V.D., Vonikakis, V., Winkler, S.: Deep learning for emotion
  recognition on small datasets using transfer learning. In: Proceedings of the
  2015 ACM on international conference on multimodal interaction. pp. 443--449
  (2015)

\bibitem{nusseck2008contribution}
Nusseck, M., Cunningham, D.W., Wallraven, C., B{\"u}lthoff, H.H.: The
  contribution of different facial regions to the recognition of conversational
  expressions. Journal of vision  \textbf{8}(8), ~1--1 (2008)

\bibitem{pham2021facial}
Pham, L., Vu, T.H., Tran, T.A.: Facial expression recognition using residual
  masking network. In: 2020 25th International Conference on Pattern
  Recognition (ICPR). pp. 4513--4519. IEEE (2021)

\bibitem{pourmirzaei2021using}
Pourmirzaei, M., Esmaili, F., Montazer, G.A.: Using self-supervised co-training
  to improve facial representation. arXiv preprint arXiv:2105.06421  (2021)

\bibitem{savchenko2021facial}
Savchenko, A.V.: Facial expression and attributes recognition based on
  multi-task learning of lightweight neural networks. arXiv preprint
  arXiv:2103.17107  (2021)

\bibitem{schiller2020relevance}
Schiller, D., Huber, T., Dietz, M., Andr{\'e}, E.: Relevance-based data
  masking: a model-agnostic transfer learning approach for facial expression
  recognition. Frontiers in Computer Science  \textbf{2}(6) (2020)

\bibitem{schoneveld2021leveraging}
Schoneveld, L., Othmani, A., Abdelkawy, H.: Leveraging recent advances in deep
  learning for audio-visual emotion recognition. Pattern Recognition Letters
  (2021)

\bibitem{selvaraju2017grad}
Selvaraju, R.R., Cogswell, M., Das, A., Vedantam, R., Parikh, D., Batra, D.:
  Grad-cam: Visual explanations from deep networks via gradient-based
  localization. In: Proceedings of the IEEE international conference on
  computer vision. pp. 618--626 (2017)

\bibitem{vo2020pyramid}
Vo, T.H., Lee, G.S., Yang, H.J., Kim, S.H.: Pyramid with super resolution for
  in-the-wild facial expression recognition. IEEE Access  \textbf{8},
  131988--132001 (2020)

\bibitem{zhou2016learning}
Zhou, B., Khosla, A., Lapedriza, A., Oliva, A., Torralba, A.: Learning deep
  features for discriminative localization. In: Proceedings of the IEEE
  conference on computer vision and pattern recognition. pp. 2921--2929 (2016)

\end{thebibliography}
}

\end{document}